# When Does Machine Learning Beat Value Sorting? A Three-Dataset Diagnostic of Exposure-Weighted Shipment Prioritization

JIZE LI

Boston University, United States, jizel@bu.edu

Delay-risk models are usually judged by predictive accuracy. What matters in practice is narrower: with capacity to review only a few shipments, which ones should a manager check first? We evaluate whether machine learning clears a demanding no-model baseline: inspect the highest-value shipments first. Across three real supply-chain contexts—SCMS procurement, DataCo logistics, and Olist e-commerce—we use leakage-controlled rolling-origin evaluation and 1,000-sample paired bootstrap confidence intervals. Ranking by predicted delay severity times known value (M1) beats severity-only ranking in all three datasets, yet it does not generally beat value sorting. At a 10% review budget, M1 minus VALUE_ONLY is −5.5 percentage points (pp) for SCMS, +10.1 pp for DataCo, and −4.9 pp for Olist. The divide is consistent with severity learnability: DataCo has $R^2 = 0.27$ and calibration bias +0.01 days, whereas SCMS and Olist have $R^2 \approx -0.02$ and negative bias. Nested-CV cost-sensitive retraining does not deliver a stable improvement over M1. The paper presents a deployment diagnostic and evaluation protocol rather than a new learning algorithm. Value sorting should remain a permanent benchmark, and ML should be deployed only after severity learnability and calibration are audited and it clears that gate in rolling-origin evaluation.

## 1 INTRODUCTION

Delivery-delay prediction becomes operationally useful only when it changes a constrained decision. A manager may have capacity to review, expedite, or escalate only the top $k$% of upcoming shipments. The practical objective is therefore not simply to classify late deliveries accurately. It is to use a limited intervention budget to capture as much delay-related business exposure as possible.

VALUE_ONLY is a serious operational baseline: shipment value is known before delivery, directly enters the exposure proxy, and requires no machine-learning model. A predictive prioritization system should therefore answer a basic question: does it capture more exposure than inspecting the highest-value shipments first? Beating severity-only ranking is not enough because managers could simply inspect high-value shipments first. Delivery-delay studies commonly emphasize predictive metrics such as accuracy, AUC, or F1, but an operational baseline and a rolling-origin evaluation under a fixed review budget can change the deployment conclusion [12, 15].

We study this question across three distinct contexts: procurement shipments in SCMS, logistics orders in DataCo, and e-commerce orders in Olist. We compare the simple value-only policy against predicted severity, predicted lateness probability, probability times value, predicted severity times value, and nested-CV cost-sensitive retraining. Every comparison uses rolling-origin test folds, training-fold-only preprocessing, shared model hyperparameters, and paired bootstrap uncertainty.

The answer depends on the dataset. At a 10% review budget, predicted severity times value (M1) exceeds value sorting by 10.10 pp on DataCo, with a 95% confidence interval (CI) of [9.67,10.42] pp. It does not clear the gate on SCMS (−5.46 pp, CI [−11.83,1.73] pp) or Olist (−4.87 pp, CI [−7.29, −2.87] pp). The three observed contexts are consistent with a simple diagnostic: value-weighted ML helps when severity is genuinely learnable and calibrated. DataCo has $R^2 = 0.270$ and calibration bias +0.010 days; SCMS and Olist have $R^2$ near −0.02 and negative calibration bias. Three datasets cannot establish a causal threshold, but two failures out of three is itself the practical message: ML prioritization is not automatically worth deploying.

We formulate delivery-risk prioritization as exposure-weighted ranking under a limited review budget and introduce VALUE_ONLY as a mandatory operational gate. Across three public datasets, leakage-controlled rolling-origin evidence shows that value-weighted ML clears this gate only conditionally and that nested-CV cost-sensitive retraining does not reliably improve over M1. The resulting deployment protocol audits severity learnability and calibration, then adopts ML only when paired bootstrap intervals support an improvement over value sorting.

## 2 RELATED WORK

### 2.1 Delay Prediction as Decision Support

Machine learning has been applied to supply-chain risk, shipment delay, and last-mile delivery prediction, with the promise that earlier identification focuses scarce attention on shipments likely to cause disruption. These models are typically reported with classification metrics such as AUC or F1 [3, 12, 14, 15, 17], and a smaller line of work evaluates rankings by top-$k$ capture, lift, or profit [7, 9]. But predictive discrimination is not itself an intervention policy: a decision-focused evaluation must measure the exposure captured under a fixed budget and compare against policies already available without a model. To our knowledge, no prior study systematically tests delay-prioritization ML against a no-model value-sorting policy across multiple datasets with a learnability/calibration mechanism.

### 2.2 Cost-Sensitive Learning and Expected Cost

Cost-sensitive learning recognizes that prediction errors can have unequal consequences. Elkan formalizes expected-cost decision making [6]. Our M2b_tuned weighting follows cost-proportionate example weighting [16]; we test whether it improves ranking beyond M1. If severity is calibrated, multiplying predicted severity by known shipment value is already an expected-exposure ranking rule, so retraining may add little or degrade ranking quality.

### 2.3 Boosting, Interpretation, and Ranking Evaluation

We use XGBoost, a scalable gradient-boosted tree implementation suitable for heterogeneous tabular data, and SHAP feature attributions for interpretation [4, 8]. Top-$k$ capture and lift metrics connect predictions to a constrained review decision, whereas accuracy metrics alone do not [7, 9].

The paper focuses on a diagnostic decision framework rather than a new learning algorithm: a deployment candidate must beat a strong no-model rule in rolling-origin evaluation.

## 3 PROBLEM FORMULATION

For shipment $i$, let $a_i$ be the actual delivery date and $s_i$ the scheduled or estimated delivery date. We define nonnegative delay severity in days as

$$d_i = \max(a_i - s_i, 0). \quad (1)$$

Here severity is absolute delay in calendar days; it is not weighted by service-level agreements, inventory buffers, or relative lead time, any of which a domain-specific deployment could substitute as a managerially weighted severity. Let $v_i \geq 0$ denote shipment or order value known at the ranking decision. The realized exposure proxy is

$$e_i = d_i v_i. \quad (2)$$

For a scoring rule $r$, let $T_k(r)$ denote the shipments in the top $k$% of scores. We evaluate

$$\text{Capture@}k(r) = \frac{\sum_{i \in T_k(r)} e_i}{\sum_i e_i}. \quad (3)$$

Lift@10% is Capture@10% divided by 0.10.

We compare seven rankers. RANDOM is a seeded random order. VALUE_ONLY ranks by known value $v_i$. PROB_ONLY ranks by predicted late-delivery probability. SEVERITY_ONLY (M0) ranks by predicted delay severity $\hat{d}_i$. BINARY×VALUE ranks by predicted late-delivery probability times $v_i$. M1 ranks by predicted severity times value, $\hat{d}_i v_i$. M2b_tuned ranks by a cost-sensitive retrained severity prediction times value; it uses the same severity model class and hyperparameters as M0 and M1, but applies nested-CV-selected value-based sample weights during training.

We define the *value-only gate* as a deployment criterion. M1 clears the gate if and only if the paired 95% bootstrap CI for

$$\Delta_{\text{gate}} = \text{Capture@}k(\text{M1}) - \text{Capture@}k(\text{VALUE_ONLY}) \quad (4)$$

is strictly positive. A positive point estimate without a strictly positive CI is not a passed gate.

The value-only gate has a direct operational role. Because VALUE_ONLY is available without model deployment, beating M0 alone is insufficient to justify a more complex prioritization system.

## 4 DATASETS

### 4.1 Three Operational Contexts

Table 1 summarizes three heterogeneous settings. SCMS is a procurement-oriented delivery history dataset [13]. DataCo covers logistics and order fulfillment with 180,519 rows [5]. Olist is assembled at order level from Brazilian e-commerce order-fulfillment, item, product, seller, customer, category-translation, and payment tables [10]. The three datasets are not treated as interchangeable replicas; they are used to test whether exposure-weighted ML deployment conclusions are context-dependent.

Table 1: Dataset scope after label-validity filtering.

| Dataset | Raw rows | Analysis rows | Late rows |
|---|---|---|---|
| SCMS | 10,324 | 10,324 | 1,186 |
| DataCo | 180,519 | 180,519 | 103,400 |
| Olist | 99,441 | 96,470 | 7,826 |

For Olist, the final analysis includes 96,470 delivered orders with valid purchase, estimated-delivery, and actual-delivery timestamps. The late rate is 8.11%, mean nonnegative severity is 0.775 days, and mean order value is 137.04 currency units. We retain only features known at purchase: product metadata, order value, freight value, item count, seller and customer location, payment fields, estimated delivery interval, and purchase-time seasonality. Across datasets, value is known at ranking time; actual delivery timestamps, derived labels, and other post-shipment fields are excluded from model features.

### 4.2 SCMS Selection-Bias Correction

The original SCMS pipeline required a purchase-order date and retained only 4,592 of 10,324 rows. That rule dropped 5,732 rows (55.52%) and removed all From-RDC fulfillment records. The corrected pipeline preserves all 10,324 rows with valid delay labels, including 5,404 From-RDC records (52.34%). This changes the evaluated population materially and is therefore a selection-bias correction, not merely a missing-value convenience.

### 4.3 Temporal Drift

Across six chronological blocks, late prevalence varies materially in SCMS (1.95% to 16.35%, ending at 15.65%) and Olist (4.46% to 17.21%, ending at 6.07%), while DataCo remains comparatively stable (57.04% to 58.02%); this non-stationarity motivates rolling-origin evaluation [1].

## 5 METHODOLOGY

### 5.1 Leakage-Controlled Pipeline

To handle non-stationarity, we use five outer rolling-origin folds rather than random splits, following temporally valid evaluation guidance [2]; each test fold follows its training data. Imputers and one-hot encoders are fit within the training fold; Olist also frequency-encodes high-cardinality seller, product-category, seller-state, and customer-state fields. Its leakage audit excludes actual delivery timestamp, post-purchase carrier events, delivery status, delay, severity target, and binary-target aliases. PROB_ONLY and BINARY×VALUE use an XGBoost classifier trained on the binary late label with the severity regressor's hyperparameters and leakage-controlled features.

The severity regressor is XGBoost with 180 trees, maximum depth 4, learning rate 0.05, subsample 0.85, column subsample 0.85, $L_1$ regularization 0, and $L_2$ regularization 1. M0, M1, and M2b_tuned receive the same model hyperparameters. M2b_tuned differs only by training-fold sample weight:

$$w_i(\alpha) = 1 + \alpha \frac{v_i}{\bar{v}_{\text{train}}}, \quad (5)$$

where $\alpha \in \{0,1,2,5,10\}$ is selected inside the outer training data by nested rolling-origin validation. The outer test fold never selects $\alpha$.

### 5.2 Target Transformation and Uncertainty

We compare raw and $\log(1 + d_i)$ severity targets using pooled rolling-origin validation RMSE. The stored pipeline selects the raw target for SCMS and DataCo and $\log(1 + d_i)$ for Olist. The comparison is transformation selection within the same model family, not a dataset-specific hyperparameter advantage for M2b_tuned.

All pooled test-fold ranking comparisons use 1,000 paired bootstrap resamples and random_state=42. Pairing preserves the fact that candidate rankers are evaluated on the same shipments. We report 95% percentile intervals and do not interpret an interval crossing zero as a proven null.

For each outer fold, we fit fold-local preprocessing and the standard models, select $\alpha$ by inner rolling-origin Capture@10%, refit weighted severity, score the later test fold, pool outer-test scores, and bootstrap paired capture differences. The gate clears only when CI(M1 − VALUE_ONLY) is strictly positive.

## 6 RESULTS

### 6.1 Value Weighting Improves Severity-Only Ranking

Ranking by predicted severity times value improves substantially upon ranking by severity alone. At $k = 10\%$, M1 minus M0 is +34.70 pp on SCMS (95% CI [27.38,40.97] pp), +11.08 pp on DataCo (CI [10.72,11.46] pp), and +27.11 pp on Olist (CI [21.97,32.32] pp). This is the robust positive result: known value matters at ranking time.

That result is not enough to justify ML deployment. VALUE_ONLY already exploits known value without prediction, and it is a strong comparator.

### 6.2 Does ML Clear the Value-Only Gate?

Table 2 is the centerpiece of our analysis. M1 clears the gate only on DataCo. SCMS has a negative point estimate and an interval crossing zero. Olist is more decisive: VALUE_ONLY captures 46.38% of realized exposure at a 10% budget, compared with 41.51% for M1; M1 minus VALUE_ONLY is −4.87 pp with CI [−7.29, −2.87] pp. Across the stored 5%, 10%, and 20% comparisons, M1 exceeds VALUE_ONLY only on DataCo and remains below it on SCMS and Olist. With three datasets, we do not estimate a universal cutoff or claim that calibration alone causes gate clearance.

Table 2: Three-dataset diagnostic at a 10% review budget. Gate clearance requires a strictly positive paired 95% CI for M1 minus VALUE_ONLY.

| Dataset | ρ | $R^2$ | M0 lift | Bias (days) | VALUE capture | M1 capture | M1-VALUE (95% CI) | Gate? |
|---|---|---|---|---|---|---|---|---|
| SCMS | 0.149 | −0.021 | 1.456 | −1.658 | 0.547 | 0.493 | −0.055 [−0.118, 0.017] | No |
| DataCo | 0.386 | 0.270 | 2.124 | 0.010 | 0.222 | 0.323 | 0.101 [0.097, 0.104] | Yes |
| Olist | 0.075 | −0.019 | 1.440 | −0.611 | 0.464 | 0.415 | −0.049 [−0.073, −0.029] | No |

### 6.3 Mechanism: Learnability and Calibration

The observed divide is theory-consistent. DataCo severity is meaningfully learnable: Spearman $\rho = 0.386$, $R^2 = 0.270$, and pooled calibration bias +0.010 days. SCMS and Olist are weak at planning time: SCMS has $\rho = 0.149$, $R^2 = -0.021$, and bias −1.658 days; Olist has $\rho = 0.075$, $R^2 = -0.019$, and bias −0.611 days. Figure 1 shows the decile-level calibration profiles.

Expected-cost logic explains the divide: multiplying predicted severity by value is an expected-exposure rule only when severity carries useful ordering [6, 11], so the quantity most relevant to the rank-based Capture@$k$ is the rank informativeness of predicted severity (Spearman $\rho$). DataCo's $\rho = 0.386$ contrasts with near-noise values on SCMS (0.149) and Olist (0.075). Calibration bias affects the magnitude of estimated exposure but not a monotone ranking, so recalibration alone cannot rescue M1 where rank signal is largely absent, and VALUE_ONLY remains superior. Because learnability and calibration co-vary across our three datasets, we report their joint association and do not claim to disentangle them.

Figure 1: Severity calibration by predicted-severity decile. SCMS and Olist show material underestimation, while DataCo is comparatively well calibrated.

### 6.4 Cost-Sensitive Retraining Does Not Rescue Ranking

Table 3 reports the second major result. Nested-CV cost-sensitive retraining does not outperform M1 on any dataset at $k = 10\%$. It is significantly worse on SCMS and Olist. DataCo is a precise non-improvement rather than a broad underpowered tie: M2b_tuned minus M1 is $-0.04$ pp with CI $[-0.09, 0.01]$ pp, whose half-width is only 0.05 pp. Any undetected DataCo advantage is operationally modest at this budget.

Table 3: M2b_tuned minus M1 at Capture@10%.

| Dataset | Difference (pp) | 95% CI (pp) |
|---|---|---|
| SCMS | −3.33 | [−9.21, −0.59] |
| DataCo | −0.04 | [−0.09, 0.01] |
| Olist | −3.03 | [−4.86, −1.37] |

### 6.5 Full Ranker Comparison

Table 4 reports Capture@10% for all rankers. At this budget, VALUE_ONLY is strongest on SCMS and Olist, while M1 is strongest on DataCo. Relative to M1, BINARY×VALUE is significantly higher on SCMS, lower on DataCo, and not significantly different on Olist; Table 3 reports M2b_tuned comparisons.

Table 4: Master results: Exposure-Capture@10%.

| Ranker | SCMS | DataCo | Olist |
|---|---|---|---|
| RANDOM | .079 | .100 | .097 |
| VALUE_ONLY | .547 | .222 | .464 |
| PROB_ONLY | .237 | .108 | .150 |
| M0 | .146 | .212 | .144 |
| BINARY×VALUE | .546 | .287 | .430 |
| M1 | .493 | .323 | .415 |
| M2b_tuned | .459 | .323 | .385 |

### 6.6 Operational Diagnostics

Warning lead time differs by context. Among M1-ranked positive-severity shipments at $k = 10\%$, the median lead-time analog is 108 days for SCMS, 2.00 days for DataCo, and 22.09 days for Olist. Means are 135.45, 1.88, and 22.34 days, respectively. These values characterize how much time an intervention policy might have, but they are dataset-specific analogs rather than guaranteed operational service levels.

SHAP associations differ by context. SCMS is led by PQ-first-sent date and project code; DataCo by shipment mode and planned duration; Olist by estimated delivery days and purchase month. These explain model behavior, not causality. DataCo's leading operational lever supports severity learning and M1; identifier- and calendar-led SCMS and Olist mostly reshuffle an already-good value ordering. DataCo's learnability plausibly reflects standardized shipment modes that expose a clean, low-cardinality delay signal at order time, whereas procurement (SCMS) and e-commerce (Olist) differ in delay mechanisms and value distributions; we therefore report per-dataset rather than aggregating, and treat cross-domain transfer as out of scope.

### 6.7 Binary Versus Severity Nuance

Severity regression is magnitude-aware but not universally superior: at 10%, BINARY×VALUE captures 54.57% versus 49.26% for M1 on SCMS, 28.72% versus 32.31% on DataCo, and 42.95% versus 41.51% on Olist, where the row-ranker-minus-M1 CI crosses zero. Candidate policies should therefore be compared at the deployment budget rather than selected from model-family labels.

## 7 DISCUSSION

### 7.1 Value Sorting Is a Serious Gate

Value sorting needs no model and is strong by construction: value is one factor in exposure itself. M1 must contribute severity ordering beyond what value already encodes, and our results show that it does so only when severity is learnable. The gate is deliberately conservative: it asks whether ML's operational burden is justified by a statistically supported gain under temporally realistic evaluation.

### 7.2 Implications for Intelligent Decision Making

Predictive performance is intermediate evidence; deployment creates a budgeted review queue. On DataCo, switching from value sorting to M1 lifts captured exposure at a 10% budget from 22% to 32%, so the same queue reaches almost half again as much delayed value. On SCMS, M1 captures about five points less than value sorting, forfeiting exposure that a database-column sort would catch. The asymmetry is the point: a model that helps in one setting can quietly cost capture in another, and only a value-only comparison makes that visible.

*Deployment checklist.* The results map onto a concrete checklist: (1) define exposure as severity × value and fix the review budget $k$; (2) take VALUE_ONLY capture at $k$ as the incumbent benchmark; (3) audit severity rank informativeness ($\rho$) and calibration on rolling-origin folds; (4) adopt M1 only if the paired 95% CI of (M1 − VALUE_ONLY) is strictly positive at $k$; and (5) re-audit under drift, and do not add cost-sensitive retraining without a stable nested-CV gain.

### 7.3 Why Retraining Need Not Help

Our negative result is consistent with expected-cost reasoning [6]: a calibrated severity estimate multiplied by known value is already a ranking-time expected-exposure rule. Sample weighting can distort severity ordering without contributing new information. This does not prove that every cost-sensitive learner is ineffective; it favors the simpler rule unless nested validation shows a stable advantage. This evidence concerns one cost-sensitive family, cost-proportionate example weighting with a nested-CV-selected weight, and does not preclude asymmetric losses, quantile or distributional regression, or other cost structures; these remain future work.

### 7.4 Beyond Delivery Delay

The value-only gate extends to decisions where exposure factorizes into known magnitude and uncertain risk, such as fraud screening, predictive maintenance, and collections. In each case, compare ML with sorting by known magnitude under a fixed budget and temporally valid evaluation.

## 8 THREATS TO VALIDITY AND LIMITATIONS

Three heterogeneous public datasets do not identify a universal or causal gate-clearance threshold, and retrospective data may lack richer pre-delivery signals. Severity times value is an interpretable proxy, not every downstream cost; alternative exposure definitions may change VALUE_ONLY's strength. Because the exposure proxy is multiplicative (severity times value), value sorting is strong by construction; under nonlinear or threshold delay penalties, including contractual breaches, reputational harm, or production halts in pharmaceutical or just-in-time settings, the gate verdict could differ, so our conclusion is conditional on this multiplicative cost model. SHAP explains fitted-model associations, not causal mechanisms [8]. Warning lead time is a dataset-specific analog rather than a guaranteed intervention window, and calibration and ranking performance require monitoring under drift.

## 9 CONCLUSION

Shipment-prioritization ML should beat a simple operational alternative: inspect high-value shipments first. Across SCMS, DataCo, and Olist, predicted severity times value consistently beats severity-only ranking, but it clears the value-only gate only on DataCo. The conditional pattern is consistent with learnability and calibration: severity is informative and cleanly calibrated on DataCo, while it is weak in SCMS and Olist. The tested nested-CV cost-sensitive retraining does not deliver a stable improvement over ranking-time multiplication. The deployment rule is simple:

retain VALUE_ONLY as a permanent benchmark, validate with rolling-origin folds, audit calibration, and deploy ML only when its paired confidence interval clears the gate.

### Data and Code Availability

All three datasets are publicly available (SCMS via USAID [13], DataCo via Mendeley Data [5], Olist via Kaggle [10]). Code and scripts will be released with the final publication.